\documentclass[a4paper,fleqn]{cas-dc}

\usepackage[numbers]{natbib}
\usepackage{algorithm}
\usepackage{algorithmic}
\usepackage{balance}
\usepackage{xspace}
\usepackage{amssymb,amsmath}
\usepackage{float}
\usepackage{flushend}
\usepackage{soul} 
\usepackage{adjustbox}

\def\tsc#1{\csdef{#1}{\textsc{\lowercase{#1}}\xspace}}
\tsc{WGM}
\tsc{QE}
\tsc{EP}
\tsc{PMS}
\tsc{BEC}
\tsc{DE}

\begin{document}
\let\WriteBookmarks\relax
\def\floatpagepagefraction{1}
\def\textpagefraction{.001}
\shorttitle{Multi-Day Ahead COVID-19 Cases Prediction in India with Gated Recurrent Unit Networks}
\shortauthors{Chakraborty et~al.}

\title [mode = title]{Transfer-Recursive-Ensemble Learning for Multi-Day COVID-19 Prediction in India using Recurrent Neural Networks}   

\author[1]{Debasrita Chakraborty}[orcid=0000-0001-7273-1353]
\address[1]{Machine Intelligence Unit, Indian Statistical Institute, Kolkata, India}
\ead{debasritac@gmail.com}

\author[2]{Debayan Goswami}[orcid=0000-0002-6133-6937]
\address[2]{Department of Computer Science and Engineering, Jadavpur University, Kolkata, India}
\ead{debayang.ju@gmail.com}

\author[2]{Susmita Ghosh}[orcid=0000-0002-1691-761X]
\cormark[1]
\ead{susmitaghoshju@gmail.com}

\author[1]{Ashish Ghosh}[orcid=0000-0003-1548-5576]
\ead{ash@isical.ac.in}

\author[3]{Jonathan H. Chan}
\address[3]{Innovative Cognitive Computing (IC2) Research Center, King Mongkut’s University of Technology Thonburi, Thailand}
\ead{jonathan@sit.kmutt.ac.th}

\author[4]{Lipo Wang}
\address[4]{School of Electrical and Electronic Engineering, Nanyang Technological University, Singapore}
\ead{ELPWang@ntu.edu.sg}

\cortext[cor1]{Corresponding author at: Department of Computer Science and Engineering, Jadavpur University, Kolkata, India \\Email address: susmitaghoshju@gmail.com (Author 3)}

\begin{abstract}
The COVID-19 pandemic has put a huge challenge on the Indian health infrastructure. With more and more people getting affected during the second wave, the hospitals were over-burdened, running out of supplies and oxygen.  In this scenario, prediction of the number of COVID-19 cases beforehand might have helped in the better utilization of limited resources and  supplies. This article deals with the prediction of new COVID-19 cases, new deaths and total active cases for multiple days in advance. The proposed method uses gated recurrent unit networks as the main predicting model. A study is conducted by building four models that are pre-trained on COVID-19 data from four different countries (United States of America, Brazil, Spain and Bangladesh) and are fine-tuned or retrained on India's data. Since the four countries chosen have experienced different types of infection curves, the pre-training provides a transfer learning to the models incorporating diverse situations into account. Each of the four models then gives 7-day ahead predictions using recursive learning method for the Indian test data. The final prediction comes from an ensemble of the predictions of the different models. This method with two countries, Spain and Bangladesh, is seen to achieve the best performance amongst all the combinations as well as compared to other traditional regression models.  
\end{abstract}


\maketitle

\section{Introduction}
The COVID-19 pandemic has placed India's relatively limited healthcare resources under tremendous pressure. Owing to the nation wide lockdown starting from 24 March 2020, imposed by the Indian government, the number of COVID-19 cases in the first wave were limited. However, with the opening up of the cities, its transport systems and allowance of various festivities, India has faced a very severe second wave with the peak at almost 0.4 million new cases a day. Although the peak seems to be over, the actual number of people to be affected in the coming days is very difficult to determine. There had been an exponential rise \cite{mondal2020fear} of new cases in the second wave and the third wave and as of 24 July 2022, there are close to 0.15 million active COVID-19 cases in India.

COVID-19 has disrupted demand projections, which help merchants and providers of consumer products and services determine how much to purchase or create, where to purchase goods, and how much to promote or sell. During the early stages of the pandemic, abrupt curfews and a migration to working from home prompted panic purchases of various food products and household supplies. Some things were sold out, while others remained on the shelf. Insecurity abounds today on several levels. Certain items, such as toilet paper and frozen foods, are still scarce. Food retailers are stocking seasons' worth of basic necessities rather than days' worth in order to best prepare for winter season, when there could be a return of illnesses and people are anticipated to stay at home. Amidst all this, there are speculations of what the next pandemic would be. The COVID-19 pandemic has showed each country their limitations.

According to a worldwide trend, we can simply state that our existing medical capacity cannot meet the high health demands caused by the coronavirus pandemic \cite{pandey2020}. Infectious diseases often rise to pandemic level whenever the risk factors simultaneously happen. The circumstances can affect the availability of hospital beds, ICU beds, fans, PPEs and qualified medical staff across the country. It is therefore challenging for the authorities to supply all sectors of society with the required healthcare services. Indian medical system had similarly collapsed during the second wave of COVID-19 infection due to the high hospital admission rate. 

If there was a prediction system available which could project a better estimate of the number of affected people much earlier on, then the authorities could have maintained stocks according to that. Researchers have proposed various models for predicting the COVID-19 cases. Various mathematical models have been used to predict and to understand the spread of the disease \cite{sanchez2020}. Auto-regressive Integrated Moving Average (ARIMA) has been used \cite{achterberg2020, hyndman2008, singh2020prediction} as the standard model to predict the behaviour of the infection curves in different countries. The model was able to capture the total case statistics because the number of total cases is seen to follow a standard exponential curve. On the contrary, the number of new cases each day is highly uncertain and involves a lot many variables and therefore, it is much more difficult to handle using ARIMA \cite{mandal2020, pandey2020}. Researchers \cite{singh2020prediction1} have therefore, explored the use of support vector machines to predict the daily COVID-19 cases. Recurrent neural networks (RNNs) have also been tested and have become the state of the art models for predicting the daily number of cases. Researchers \cite{shahid2020} have compared various RNN models like Long Short Term Memory (LSTM) \cite{alazab2020,hochreiter1997long,nikparvar2021}, Gated Recurrent Unit (GRU) and Bi-LSTMs. They have observed that these models are more robust than ARIMA or Support Vector Regression (SVR) \cite{Yamak2019}. Transfer learning and ensemble modelling have also been applied to study the statistics of daily COVID-19 cases \cite{debasrita2021} and it has shown to perform even better than the standard LSTM-RNNs. Ensemble learning in conjunction with transfer learning has, however, not been tested before. It might capture trends of multiple countries and take advantage of the knowledge of infection spread trends that the test country has not experienced before.

As we have witnessed during the peaks of the infection waves that the medical supplies were generally distributed to an area based on the real-time daily new cases in that particular area/ hospital. It takes some time to get these supplies and meanwhile the patients might be critical \cite{moran2016messier}. Therefore, a predictive method for the estimation of COVID-19 statistics for multiple days in advance, can provide a better framework for medical logistics. A predictive model in the similar direction has been proposed in this article. Some researchers \cite{fritz2022, rajesh2020, ribeiro2020short} have attempted the multi-step prediction of cumulative COVID-19 cases. However, multi-step prediction of daily new cases, daily fatalities and total active cases simultaneously is difficult due to their chaotic nature. The proposed model alleviates these problems and gives a multi-variable (3 COVID-19 parameters: New Cases, New Deaths and Active Cases) multi-day prediction for the daily statistics.

The proposed method draws motivation from the concept of recursive learning. The aim of recursive learning is to establish a model that can learn to fill in the missing parts in its input. In prediction tasks, the recursive model trains on its predictions \cite{yoo1999short}. This provides the input with a feedback from the output. The proposed method has been tested to give predictions for seven days ahead case. To achieve this, the proposed method uses a combination of several learning methodologies and integrates them into one. It has been shown that such a combined model is more efficient at providing multi-day forecasts than the existing standard models.

The contribution of this work can be highlighted along the following lines:
\begin{itemize}
  \item Introduction of a transfer learning scenario for incorporating COVID-19 spread behavior in different countries .
  \item Recursive learning for 7-day prediction i.e., using the predictions recursively for new predictions.
  \item Combination of the predictions from different models using a weighted ensemble.
\end{itemize}

Rest of the article is organised as follows: The preliminaries for the proposed method is outlined in details in Section 2 with special emphasis on transfer learning and recursive learning employed. Section 3 describes the proposed method. Section 4 contains the experimental results, along with a comparison with existing standard methods. Section 5 contains a discussion of the results, along with statistical significance testing of the models. Section 6 concludes this article with the scope for future research.

\section{Preliminaries}

This section describes the dataset and the basic building blocks which have been used in the proposed model.

\subsection{Dataset}

The data for this study has been taken from the database of Worldometers website \cite{Worldometer}. This website provides COVID-19 related data including new cases, new deaths, active cases, total tests etc. for 222 different countries all along the period of the pandemic. To predict the multi-day ahead COVID-19 cases in India we have considered data related to daily new cases, daily new deaths, and active cases for six different countries: the USA, Brazil, Spain, Bangladesh, Australia and India for the period from 15 February 2020 to 16 June 2022.

\subsection{Recurrent Neural Networks}

The current solution relies on gated recurrent units (GRUs) as the basic building blocks for multi-day prediction of COVID-19 parameters. Gated recurrent units and long short-term memory (LSTMs), as prevalent members of recurrent neural networks (RNNs), have the innate ability to capture trends and seasonality in time-series data \cite{Alex2020}. They are, therefore, the go-to methods for time series predictions.

GRUs are able to solve the exploding and vanishing gradient problem that is common for vanilla RNNs \cite{cho2014}. With its reset gate and update gate, a GRU is able to decide what information to keep from the previous state and what information to pass on to the next state. This gives GRUs the ability to keep relevant information from much earlier in the sequence while removing information that is no longer relevant for the task at hand. For a more detailed description of GRU, one can refer to \cite{cho2014}.

GRUs are one of the simplest recurrent neural network models and have been used for time series prediction tasks in multiple domains, e.g., for traffic flow prediction \cite{Fu2016}, energy load forecasting \cite{Kumar2018}, stock market forecasting \cite{Althelaya2018}, air pollution forecasting \cite{Tao2019}. It has also been used for COVID-19 prediction with the help of deep learning based models \cite{shahid2020}.

The present work of multi-day COVID-19 prediction has been done using GRUs as the basic building blocks in order to harness its effective sequence modelling function and also to prevent over-fit in our relatively small dataset.

LSTMs are also used for time-series prediction in multiple domains \cite{gers2002, karevan2020, li2019, yadav2020}. In the present work, we have compared the performances of GRUs with that of LSTMs, considering both as the basic building blocks.

\subsection{Transfer Learning}

Transfer learning is the scenario where a pre-trained model for one particular problem is applied to a second, different but related problem \cite{torrey2010}. Transfer learning tries to take advantage of what has already been learned in a problem and applies it to improve the generalization in another related problem.

The domain in which the model is trained is called the source domain and the domain in which the model is applied is called the target domain \cite{pan2009}. The source and the target domain may be different enough but need to have some sort of a relation. The predictive model in the source domain needs to be similar to that of target domain in order for transfer learning to work. Transfer learning is mainly applied in such target domains where sufficient labeled data is not available \cite{weiss2016}.

Transfer learning has been applied in the COVID-19 scenario for different tasks. It has been used for classification of COVID-19 from non-COVID-19 patients by using chest CT images \cite{pathak2020}, for face-mask detection in public areas \cite{horry2020}, COVID-19 cases and death forecasts using LSTMs \cite{gautam2021} etc. 

In the present work, transfer learning has been chosen for the task of COVID-19 case prediction to learn from the experiences of countries affected by COVID-19. Countries with different circumstances, different climates, different measures for infection control are chosen as the source domain and COVID-19 cases prediction for India is done as the target domain. In one of our previous works \cite{debasrita2021}, it is seen that transfer learning has given better results for next day prediction for COVID-19 cases using LSTMs. In this present work, we are exploiting transfer learning with recursion for multi-day ahead prediction. The details are given below.


\subsection{Recursive Learning}

\begin{figure}[h!]
    \centering
    \includegraphics[scale=0.45]{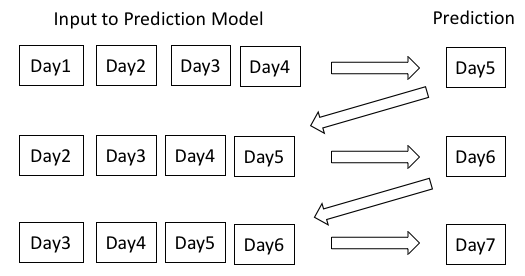}
    \caption{Recursive learning used in COVID-19 prediction}
    \label{fig:recursive_learning}
\end{figure}

The GRU model built in this work is able to predict the next day parameter after looking at the parameters over a period of past days (called the look-back period). As mentioned, in order to achieve a multi-day prediction of COVID-19 cases, a recursive learning methodology is adopted as shown in Figure \ref{fig:recursive_learning}.

In a recursive way, the predicted output of the model is fed back to the input in the next step to obtain the subsequent prediction. As an example, in Figure \ref{fig:recursive_learning}, data for day 1 to day 4 is used as input to the model to predict the data for day 5. In the next step, the data for day 5 is added to the input and the data for day 2 to day 5 is taken as input to the model to predict the output for day 6. This process is repeated recursively till the required days of prediction is obtained.

This recursive learning methodology uses the sliding window approach with the previous predictions being used as a part of the input to do future predictions. It is intuitive that the performance will vary with the change in look-back period.

In this work, this recursive learning methodology works for 7 steps to predict the COVID-19 cases for the next 7-days. However, this process of recursion can be used for prediction of COVID-19 cases for any number of days in advance.

\subsection{Ensemble Learning}

Through ensemble learning one could exploit the unique abilities of multiple models in an integrated manner by combining the results obtained from various models. In this work, the results obtained through recursive learning approach initiated from various transfer-learnt models (trained on data from respective countries) are ensembled to obtain final predictions. Several ensemble techniques exist in literature \cite{dietterich2000, zhang2012}. In the present approach, we have proposed a weighted ensemble technique. 

\subsection{Performance Metric}
To assess the performance of the proposed approach, two metrics have been used for this study. The first is the relative mean squared error (R-MSE) and the second one is relative mean absolute error (R-MAE). Rather than considering actual value, calculating the error as a fraction of the actual value is seen to be effective. As the error value is compared relative to the actual value of the parameter, hence the term relative has been used for the standard error metrics of mean squared error and mean absolute error.

These metrics are defined as follows:

\begin{multline}
Relative-MSE (R-MSE) = \\
\frac{\sum_{i=1}^{d} \left (\frac{original(i)-predicted(i)}{original(i)} \right) ^ 2}{d}
\label{eq1}
\end{multline}

\begin{multline}
Relative-MAE (R-MAE) = \\
\frac{\sum_{i=1}^{d}\left|\dfrac{original(i)-predicted(i)}{original(i)}\right|}{d}
\label{eq2}
\end{multline}

where, $predicted(i)$  is the predicted value on the $i^{th}$ day and $original(i)$  is the actual value on the $i^{th}$ day. $d$ is the total number of days involved. Note that, R-MSE better reflects the error for higher deviations than that of R-MAE as it penalizes higher deviations with a greater error value.

For each of the three COVID-19 parameters predicted (daily new cases, daily new deaths and total active cases), these two errors values (as an average of all the prediction errors over the test set) have been shown in the results section. 

\section{Proposed Method}

The proposed model is an ensemble of four different models pre-trained on data from four different countries (The United States of America (USA), Brazil, Spain and Bangladesh) in order to predict the COVID-19 daily new cases, daily new deaths and active cases for India for the next 7-days. The idea was to learn the infection spread in the worst affected country from each of the seven continents. 

The USA and Brazil were obvious choices from the North American and South American continents due to high number of COVID-19 infections. The USA has witnessed the most number of cases. The first wave in the USA was prolonged and the second wave has resulted in an increased number of deaths per day. The third and fourth waves also have a similar pattern. The USA has not witnessed a plateau in the total number of cases after the first wave. However, the daily new cases have decreased considerably after the peak of the third wave. Brazil has the third highest number of cases and currently the death rate is 3,015 per million population, which is one of the highest in the world. However, the total cases curve in Brazil has a prolonged second wave and severe third wave. Spain has been chosen from the European continent due to its well marked first, second and third waves of infection as compared to the initial plateau of infections in Italy. Bangladesh has been chosen from the Asian continent to incorporate similar climatic conditions and being a neighbouring country of the test country, India. South Africa was first taken into consideration from the African continent, however it was not incorporated into the model as the number of cases in the waves of infection have been quite low as compared to the other countries selected. Australia was also taken into consideration, but was not introduced in the model due to the very late nature of infection with it still being in the first wave of infections. To incorporate population density into account, the data of each country is divided by the corresponding population density.

\begin{figure}
  \includegraphics[width=\linewidth]{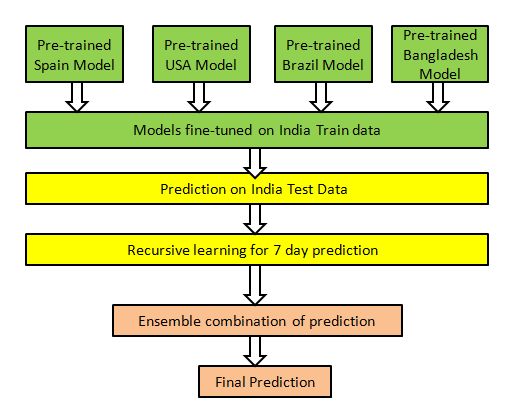}
  \caption{Flowchart of the proposed method}
  \label{fig:proposed_methd_flowchart}
\end{figure}

India witnessed a decline in the number of daily cases which suggested the ending of the third wave. However, cases have started increasing in some parts of the country again, signalling a possible fourth wave. This obviously brings us to the crucial question about the condition in India about the subsequent waves of infection. Now, since all of the four countries have shown different trends, it is not sure as to which path Indian trend would follow. This is why all possible combinations of these four countries were taken into account. More countries could have been taken for pre-training, but that would have added more complexity to the model. Training of each of these models has been done using a sliding window technique with a look back period of 14 days. This look back period has been varied from 7 to 19 in order to find the best look back period for the 7-day ahead prediction task. The proposed model is shown in Figure \ref{fig:proposed_methd_flowchart}. It consists of three main steps which are discussed in the subsequent sections.

\subsection{Step 1: Train-Test Splitting of Input Data}

The data from the period 15 February 2020 to 31 December 2021 has been used for training the models for the individual four countries. This period has been chosen to take 80\% of the total data for training. Indian data for the period 15 February 2021 to 31 December 2021 has been used for fine-tuning the transfer learning models before testing them on Indian data for the period 16 January 2022 to 16 June 2022. The remaining period of 01 January 2022 to 15 January 2022 for the Indian data has been used for cross-validation in the ensemble weighted averaging as mentioned in the subsequent paragraphs.

\subsection{Step 2: Forming the Model}

Two RNN models (LSTMs and GRUs)  have been taken independently as the building blocks for the proposed method. The proposed RNN models with parameters and their values chosen are given in Table \ref{tab:RNN_parameters}. As mentioned earlier, the GRU and the LSTM models are trained and tested in order to do performance comparison of the two RNN types.

\begin{table}[h!]
\centering
\begin{tabular}{@{}ll@{}}
\toprule
\textbf{Parameter} & \textbf{Values} \\ \midrule
Layers             & 3               \\
Number of Neurons  & {30,80,50}      \\
Learning Rate      & 0.001        \\
Optimizer          & Adam            \\ \bottomrule
\end{tabular}
\caption{Values of the parameters of the RNN models}
\label{tab:RNN_parameters}
\end{table}

\subsection{Step 3: Transfer Learning}

As stated earlier, the proposed model consists of all sixteen possible combinations of four RNN networks of two types (GRUs and LSTMs), each of which is pre-trained on data from four different countries. To pre-train each model, the data of each of the countries is taken from 15 February 2020 to 31 December 2021, as mentioned earlier. The models built on the individual countries need to be fine-tuned on Indian data in order to take into account the recent trend of COVID-19 infections in the target country, India. Therefore, the pre-trained models have been fine-tuned on Indian data for the same period 15 February 2020 to 31 December 2021. 

This method of pre-training followed by fine-tuning introduces a transfer learning \cite{yang2020} ability in the individual models. Li et. al. \cite{li2021alert} has shown that transfer learning can improve forecasting models that are based on deep learning. They have built a source domain of 12 countries, combining their data, and tried to predict the confirmed cases per million for the target countries. However, the scope of the study is limited by the prediction of just one COVID-19 parameter and also being tested on a shorter time period (31/12/2019 to 31/05/2020). Whereas, in the proposed approach, three COVID-19 parameters are predicted over an advance period of 7 days using an ensembled approach of transfer-learnt recursive model on a larger time period.

\subsection{Step 4: Recursive Learning}

To obtain 7-day ahead predictions, we have incorporated recursive learning in the proposed model. If the COVID-19 parameters for the next 7-days is to be predicted at the ${n}^{th}$ day, in the first step, the COVID-19 parameters for the next day (${(n+1)}^{th}$ day) will be predicted by the model. This prediction for the ${(n+1)}^{th}$ day is then fed back to the input to make the new input frame for predicting the COVID-19 parameters for the ${(n+2)}^{th}$ day. This process is repeated 7 times to get the predicted values for the 7-days (${(n+1)}^{th}$ day to ${(n+7)}^{th}$ day). An example of the process involved is shown in the Figure \ref{fig:recursive_learning}. This method can be applied to predict the COVID-19 parameters for a different time period by changing the period setting. However, to do a stable comparison, the prediction period has been set at 7-days for this study. Experiments are also done to study the effect of varying the look-back period used for prediction.

\subsection{Step 5: Ensemble}
Once the predictions are obtained for subsequent 7-days, the predictions from the combination of models are aggregated using weighted averaging. The weights are calculated based on the relative mean squared error (R-MSE) obtained from model cross-validation on validation data. 15 days of data for India (01 January 2022 to 15 January 2022) is kept aside for this validation task.

The relative mean squared error (R-MSE) for the validation data is calculated using the Eqs. \ref{eq1}. The weights ($w_i$) for the Model $i$ are given by Eq. \ref{eq3}, and the final prediction on any date D is given by Eq. \ref{eq4}. We have also compared our proposed ensemble method with an equally weighted ensemble technique.

\begin{equation}
w_i=\dfrac{1/R-MSE(Model~i)}{\sum_{j=1}^{n}1/R-MSE(Model~j)},
\label{eq3}
\end{equation}

where, $R-MSE(Model~i)$ is the relative $mse$ obtained for the $i^{th}$ model on validation data, $n$ is the number of models involved in the ensemble. The transfer learnt model with less error is given a higher weight as per Eq. \ref{eq3}.

\begin{equation}
prediction(D)=\sum_{i=1}^{n} w_i*prediction_i(D),
\label{eq4}
\end{equation}
where, $prediction_i(D)$ is the prediction by the $i^{th}$ model for date D. This is a weighted summation of the predictions from the transfer learnt models. 

\section{Experimental Results}

In order to predict the cases on any date, we need to use a look-back period. This period, in this case, is the number of days our model looks at for doing the predicting task. Finding the optimum value of the look-back period is crucial for the proposed method. This is because, depending on the look-back period, the performance of the models varies rapidly. 

It is to be noted that, since we are relying on a recursive learning based multi-day prediction, where the predicted values are used as inputs for the subsequent predictions, we cannot afford to take the look-back to be smaller than the number of days in the multi-day prediction task. This will result in the last few predictions (of the recursive learning methodology) being made only on predictions and not on any actual data. We have experimented with a wide range of look-back periods and a value of 14 gives the best results for all the three variables.

The values of R-MSE and R-MAE (averaged over 20 runs) obtained for all 16 combinations using the proposed model (using both GRUs and LSTMs) with those for the support vector regression (SVR), auto-regressive integrated moving average (ARIMA) and Facebook Prophet models are shown in the Tables \ref{tab:results_finetune} and \ref{tab:results_lstm}. The results are shown separately for the 3 COVID-19 parameters predicted in this study (new cases, new deaths and active cases). For establishing the efficacy of fine-tuning on Indian data, results obtained for the above mentioned methods, using GRUs, without fine-tuning are also put in Table \ref{tab:results_nofinetune}. The results put in this table are the average of 20 independent runs. It is seen that the combination model of Spain and Bangladesh gives the best results for multi-day forecasting of all the three predicted variables.

\begin{table*}[]
\begin{adjustbox}{width=1\textwidth}
\begin{tabular}{@{}lllllllllllll@{}}
\toprule
\multicolumn{1}{c}{} & \multicolumn{4}{c}{New Cases} & \multicolumn{4}{c}{New Deaths} & \multicolumn{4}{c}{Active Cases} \\ \cmidrule(l){2-13} 
\multicolumn{1}{c}{} & \multicolumn{2}{c}{R-MSE} & \multicolumn{2}{c}{R-MAE} & \multicolumn{2}{c}{R-MSE} & \multicolumn{2}{c}{R-MAE} & \multicolumn{2}{c}{R-MSE} & \multicolumn{2}{c}{R-MAE} \\ \cmidrule(l){2-13}
\multicolumn{1}{c}{\multirow{-3}{*}{Method}} & \multicolumn{1}{c}{Mean} & \multicolumn{1}{c}{STD} & \multicolumn{1}{c}{Mean} & \multicolumn{1}{c}{STD} & \multicolumn{1}{c}{Mean} & \multicolumn{1}{c}{STD} & \multicolumn{1}{c}{Mean} & \multicolumn{1}{c}{STD} & \multicolumn{1}{c}{Mean} & \multicolumn{1}{c}{STD} & \multicolumn{1}{c}{Mean} & \multicolumn{1}{c}{STD} \\ \cmidrule(r){1-13}
Spain & 0.0043 & 0.0025 & 0.0586 & 0.0224 & 0.0048 & 0.0032 & 0.0654 & 0.0283 & 0.0016 & 0.001 & 0.0258 & 0.0189 \\
USA & 0.0052 & 0.0027 & 0.0709 & 0.0188 & 0.0065 & 0.0018 & 0.0886 & 0.0166 & 0.0015 & 0.001 & 0.0209 & 0.015 \\
Brazil & 0.0038 & 0.0013 & 0.0528 & 0.0231 & 0.0032 & 0.0012 & 0.0436 & 0.0137 & 0.001 & 0.0004 & 0.0174 & 0.0115 \\
Bangladesh & 0.0035 & 0.0024 & 0.0477 & 0.0262 & 0.0037 & 0.0025 & 0.0504 & 0.0212 & 0.0018 & 0.0013 & 0.0265 & 0.0198 \\
\rowcolor[HTML]{FFFF00} 
Spain-Bangladesh & 0.0013 & 0.0009 & 0.0177 & 0.0127 & 0.0021 & 0.0011 & 0.0286 & 0.0249 & 0.0009 & 0.0003 & 0.017 & 0.011 \\
Brazil-Bangladesh & 0.0025 & 0.0016 & 0.035 & 0.015 & 0.0033 & 0.0012 & 0.0450 & 0.0138 & 0.0011 & 0.0005 & 0.0189 & 0.0128 \\
Spain-Brazil & 0.0028 & 0.0014 & 0.0382 & 0.0144 & 0.0026 & 0.0013 & 0.0357 & 0.0148 & 0.001 & 0.0007 & 0.0189 & 0.0126 \\
Spain-USA & 0.0032 & 0.0023 & 0.0442 & 0.0194 & 0.0029 & 0.0021 & 0.0395 & 0.0203 & 0.0014 & 0.001 & 0.0221 & 0.0159 \\
USA-Brazil & 0.0038 & 0.0015 & 0.0518 & 0.0139 & 0.0038 & 0.0012 & 0.0518 & 0.0131 & 0.001 & 0.0008 & 0.0176 & 0.0117 \\
USA-Bangladesh & 0.0026 & 0.0017 & 0.0354 & 0.0201 & 0.0034 & 0.0018 & 0.0463 & 0.0171 & 0.0015 & 0.001 & 0.0222 & 0.0157 \\
Brazil-USA-Bangladesh & 0.0027 & 0.0017 & 0.0368 & 0.0154 & 0.0036 & 0.0012 & 0.0491 & 0.0137 & 0.0011 & 0.001 & 0.0187 & 0.0127 \\
Spain-Brazil-Bangladesh & 0.0016 & 0.0009 & 0.0218 & 0.0162 & 0.0027 & 0.0013 & 0.0374 & 0.0156 & 0.0011 & 0.0008 & 0.0199 & 0.0137 \\
Spain-Brazil-USA & 0.002 & 0.001 & 0.0273 & 0.0151 & 0.0028 & 0.0013 & 0.0382 & 0.0147 & 0.0011 & 0.001 & 0.0188 & 0.0128 \\
Spain-USA-Bangladesh & 0.0019 & 0.0011 & 0.0259 & 0.0205 & 0.0025 & 0.002 & 0.0341 & 0.0202 & 0.0014 & 0.0008 & 0.0228 & 0.0165 \\
USA-Brazil-Bangladesh-Spain & 0.002 & 0.0013 & 0.0273 & 0.0163 & 0.0025 & 0.0013 & 0.0348 & 0.0154 & 0.0011 & 0.0006 & 0.0197 & 0.0135 \\
USA-Brazil-Bangladesh-Spain - Australia & 0.0024 & 0.0014 & 0.0312 & 0.0243 & 0.0029 & 0.002 & 0.0388 & 0.0285 & 0.0015 & 0.0006 & 0.0231 & 0.0269 \\
India & 0.0037 & 0.0021 & 0.0519 & 0.0253 & 0.0042 & 0.0034 & 0.0492 & 0.0310 & 0.0014 & 0.001 & 0.0233 & 0.0192 \\
SVR (RBF) & \multicolumn{1}{r}{1.4627} & \multicolumn{1}{r}{NA} & \multicolumn{1}{r}{1.1538} & \multicolumn{1}{r}{NA} & \multicolumn{1}{r}{0.1351} & \multicolumn{1}{r}{NA} & \multicolumn{1}{r}{0.4286} & \multicolumn{1}{r}{NA} & \multicolumn{1}{r}{0.7528} & \multicolumn{1}{r}{NA} & \multicolumn{1}{r}{0.9521} & \multicolumn{1}{r}{NA} \\
SVR (Poly) & \multicolumn{1}{r}{0.8453} & \multicolumn{1}{r}{NA} & \multicolumn{1}{r}{0.9521} & \multicolumn{1}{r}{NA} & \multicolumn{1}{r}{0.0948} & \multicolumn{1}{r}{NA} & \multicolumn{1}{r}{0.2794} & \multicolumn{1}{r}{NA} & \multicolumn{1}{r}{0.4821} & \multicolumn{1}{r}{NA} & \multicolumn{1}{r}{0.7364} & \multicolumn{1}{r}{NA} \\
ARIMA & \multicolumn{1}{r}{0.2351} & \multicolumn{1}{r}{NA} & \multicolumn{1}{r}{0.4928} & \multicolumn{1}{r}{NA} & \multicolumn{1}{r}{0.0537} & \multicolumn{1}{r}{NA} & \multicolumn{1}{r}{0.2263} & \multicolumn{1}{r}{NA} & \multicolumn{1}{r}{0.1302} & \multicolumn{1}{r}{NA} & \multicolumn{1}{r}{0.4826} & \multicolumn{1}{r}{NA} \\
Facebook Prophet & \multicolumn{1}{r}{0.0673} & \multicolumn{1}{r}{NA} & \multicolumn{1}{r}{0.2972} & \multicolumn{1}{r}{NA} & \multicolumn{1}{r}{0.0185} & \multicolumn{1}{r}{NA} & \multicolumn{1}{r}{0.1492} & \multicolumn{1}{r}{NA} & \multicolumn{1}{r}{0.0464} & \multicolumn{1}{r}{NA} & \multicolumn{1}{r}{0.2273} & \multicolumn{1}{r}{NA} \\ \bottomrule
\end{tabular}
\end{adjustbox}
\caption{Comparison of results i.e. Mean and Standard Deviation(STD) of 20 runs, from all possible combinations of four countries considered in this study. The GRU models have been fine-tuned on Indian data. Results are also shown for SVR, ARIMA and Facebook Prophet models.}
\label{tab:results_finetune}
\end{table*}

\begin{table*}[]
\begin{adjustbox}{width=1\textwidth}
\begin{tabular}{@{}lllllllllllll@{}}
\toprule
\multicolumn{1}{c}{} & \multicolumn{4}{c}{New Cases} & \multicolumn{4}{c}{New Deaths} & \multicolumn{4}{c}{Active Cases} \\ \cmidrule(l){2-13} 
\multicolumn{1}{c}{} & \multicolumn{2}{c}{R-MSE} & \multicolumn{2}{c}{R-MAE} & \multicolumn{2}{c}{R-MSE} & \multicolumn{2}{c}{R-MAE} & \multicolumn{2}{c}{R-MSE} & \multicolumn{2}{c}{R-MAE} \\ \cmidrule(l){2-13} 
\multicolumn{1}{c}{\multirow{-3}{*}{Method}} & Mean & STD & Mean & STD & Mean & STD & Mean & STD & Mean & STD & Mean & STD \\ \cmidrule(r){1-13}
Spain & 0.0042 & 0.0038 & 0.0584 & 0.0263 & 0.0063 & 0.0031 & 0.0665 & 0.027 & 0.0024 & 0.0011 & 0.0268 & 0.0151 \\
USA & 0.0054 & 0.0035 & 0.0710 & 0.0201 & 0.0064 & 0.0023 & 0.0889 & 0.015 & 0.0009 & 0.00042 & 0.0201 & 0.0162 \\
Brazil & 0.0041 & 0.0028 & 0.0526 & 0.0328 & 0.0030 & 0.0021 & 0.0440 & 0.017 & 0.0014 & 0.001 & 0.0186 & 0.0151 \\
Bangladesh & 0.0034 & 0.0023 & 0.0476 & 0.0317 & 0.0049 & 0.0015 & 0.0513 & 0.026 & 0.0016 & 0.0005 & 0.0260 & 0.019 \\
\rowcolor[HTML]{FFFF00} 
Spain-Bangladesh & 0.0015 & 0.0012 & 0.0178 & 0.0129 & 0.0027 & 0.0017 & 0.0299 & 0.017 & 0.0013 & 0.0005 & 0.0176 & 0.012 \\
Brazil-Bangladesh & 0.0027 & 0.0018 & 0.0348 & 0.012 & 0.0046 & 0.0029 & 0.0461 & 0.0217 & 0.0015 & 0.0007 & 0.0191 & 0.0132 \\
Spain-Brazil & 0.0028 & 0.0013 & 0.0383 & 0.0156 & 0.0027 & 0.0016 & 0.0347 & 0.0174 & 0.0016 & 0.0009 & 0.0203 & 0.0171 \\
Spain-USA & 0.0029 & 0.0016 & 0.0441 & 0.0173 & 0.0040 & 0.0023 & 0.0400 & 0.0214 & 0.0014 & 0.0009 & 0.0219 & 0.0153 \\
USA-Brazil & 0.0040 & 0.0024 & 0.0518 & 0.0169 & 0.0045 & 0.0028 & 0.0528 & 0.0142 & 0.0020 & 0.00064 & 0.0190 & 0.0141 \\
USA-Bangladesh & 0.0025 & 0.002 & 0.0354 & 0.0263 & 0.0036 & 0.0019 & 0.0469 & 0.0325 & 0.0026 & 0.0009 & 0.0241 & 0.0182 \\
Brazil-USA-Bangladesh & 0.0027 & 0.0018 & 0.0370 & 0.0157 & 0.0042 & 0.0016 & 0.0505 & 0.0229 & 0.0016 & 0.0008 & 0.0195 & 0.0133 \\
Spain-Brazil-Bangladesh & 0.0017 & 0.0011 & 0.0219 & 0.0131 & 0.0035 & 0.0028 & 0.0361 & 0.0151 & 0.0021 & 0.0006 & 0.0213 & 0.0188 \\
Spain-Brazil-USA & 0.0017 & 0.0011 & 0.0272 & 0.0183 & 0.0027 & 0.0019 & 0.0388 & 0.0216 & 0.0020 & 0.0011 & 0.0200 & 0.014 \\
Spain-USA-Bangladesh & 0.0020 & 0.0017 & 0.0258 & 0.0193 & 0.0039 & 0.0028 & 0.0357 & 0.0317 & 0.0026 & 0.0015 & 0.0236 & 0.0158 \\
USA-Brazil-Bangladesh-Spain & 0.0023 & 0.0015 & 0.0275 & 0.0181 & 0.0027 & 0.0015 & 0.0363 & 0.0273 & 0.0015 & 0.0009 & 0.0209 & 0.013 \\
USA-Brazil-Bangladesh-Spain - Australia & 0.0025 & 0.0014 & 0.0311 & 0.0212 & 0.003 & 0.0017 & 0.0421 & 0.0253 & 0.0019 & 0.0007 & 0.0292 & 0.0192 \\
India & 0.0035 & 0.0033 & 0.0427 & 0.0323 & 0.0047 & 0.0029 & 0.0535 & 0.0382 & 0.0022 & 0.0013 & 0.0225 & 0.0120 \\ \bottomrule
\end{tabular}
\end{adjustbox}
\caption{Comparison of results i.e. Mean and Standard Deviation(STD) of 20 runs, from all possible combinations of four countries considered in this study. Similar configuration LSTM models, fine-tuned on Indian data, have been used in this case.}
\label{tab:results_lstm}
\end{table*}

\begin{table*}[]
\begin{adjustbox}{width=1\textwidth}
\begin{tabular}{@{}lllllllllllll@{}}
\toprule
\multicolumn{1}{c}{} & \multicolumn{4}{c}{New Cases} & \multicolumn{4}{c}{New Deaths} & \multicolumn{4}{c}{Active Cases} \\ \cmidrule(l){2-13} 
\multicolumn{1}{c}{} & \multicolumn{2}{c}{R-MSE} & \multicolumn{2}{c}{R-MAE} & \multicolumn{2}{c}{R-MSE} & \multicolumn{2}{c}{R-MAE} & \multicolumn{2}{c}{R-MSE} & \multicolumn{2}{c}{R-MAE} \\ \cmidrule(l){2-13} 
\multicolumn{1}{c}{\multirow{-3}{*}{Method}} & Mean & STD & Mean & STD & Mean & STD & Mean & STD & Mean & STD & Mean & STD \\ \cmidrule(r){1-13}
Spain & 0.0065 & 0.0035 & 0.1806 & 0.0254 & 0.0102 & 0.0026 & 0.2010 & 0.034 & 0.0102 & 0.001 & 0.2007 & 0.0194 \\
USA & 0.0071 & 0.0047 & 0.1943 & 0.018 & 0.011 & 0.0015 & 0.2149 & 0.018 & 0.0117 & 0.001 & 0.2182 & 0.023 \\
Brazil & 0.0042 & 0.0023 & 0.1248 & 0.0235 & 0.0099 & 0.0019 & 0.1685 & 0.0194 & 0.0088 & 0.0008 & 0.1628 & 0.0182 \\
Bangladesh & 0.0039 & 0.0014 & 0.1824 & 0.0241 & 0.0096 & 0.0024 & 0.2180 & 0.028 & 0.0107 & 0.0017 & 0.2232 & 0.021 \\
\rowcolor[HTML]{FFFF00} 
Spain-Bangladesh & 0.0021 & 0.0013 & 0.0758 & 0.0152 & 0.0044 & 0.001 & 0.1203 & 0.027 & 0.0074 & 0.0007 & 0.1287 & 0.014 \\
Brazil-Bangladesh & 0.0028 & 0.0015 & 0.1229 & 0.018 & 0.0068 & 0.0032 & 0.1495 & 0.0149 & 0.0097 & 0.0005 & 0.1652 & 0.0159 \\
Spain-Brazil & 0.0031 & 0.0017 & 0.1957 & 0.0123 & 0.0065 & 0.0019 & 0.2036 & 0.0193 & 0.0099 & 0.001 & 0.2222 & 0.0292 \\
Spain-USA & 0.0035 & 0.0026 & 0.1592 & 0.0182 & 0.0075 & 0.0028 & 0.2026 & 0.0321 & 0.0077 & 0.0009 & 0.2037 & 0.0172 \\
USA-Brazil & 0.0037 & 0.0013 & 0.2608 & 0.0157 & 0.0067 & 0.0015 & 0.2749 & 0.0184 & 0.0111 & 0.00072 & 0.2281 & 0.0315 \\
USA-Bangladesh & 0.0032 & 0.0027 & 0.2166 & 0.0232 & 0.0073 & 0.0021 & 0.2594 & 0.0452 & 0.0086 & 0.002 & 0.2485 & 0.0273 \\
Brazil-USA-Bangladesh & 0.003 & 0.0023 & 0.0948 & 0.0192 & 0.006 & 0.0014 & 0.1715 & 0.0241 & 0.0080 & 0.003 & 0.1894 & 0.0153 \\
Spain-Brazil-Bangladesh & 0.0022 & 0.0012 & 0.0969 & 0.0185 & 0.0057 & 0.0017 & 0.1375 & 0.0174 & 0.0076 & 0.0001 & 0.1409 & 0.0142 \\
Spain-Brazil-USA & 0.0029 & 0.0015 & 0.2039 & 0.011 & 0.0082 & 0.0017 & 0.2316 & 0.0182 & 0.0094 & 0.0013 & 0.2377 & 0.0182 \\
Spain-USA-Bangladesh & 0.0028 & 0.0014 & 0.0979 & 0.032 & 0.005 & 0.0024 & 0.1277 & 0.0328 & 0.0085 & 0.0009 & 0.1489 & 0.0172 \\
USA-Brazil-Bangladesh-Spain & 0.0024 & 0.0016 & 0.0990 & 0.0249 & 0.0052 & 0.0018 & 0.1321 & 0.0182 & 0.0083 & 0.0007 & 0.1511 & 0.0142 \\
USA-Brazil-Bangladesh-Spain - Australia & 0.0029 & 0.0018 & 0.1820 & 0.038 & 0.0064 & 0.0025 & 0.1585 & 0.0271 & 0.0093 & 0.0006 & 0.1709 & 0.0249 \\ \bottomrule
\end{tabular}
\end{adjustbox}
\caption{Comparison of results i.e. i.e. Mean and Standard Deviation(STD) of 20 runs, from all possible combinations of four countries considered in this study. The GRU models have not been fine-tuned on Indian data.}
\label{tab:results_nofinetune}
\end{table*}

\section{Discussion}

It is clearly visible from Tables \ref{tab:results_finetune} and \ref{tab:results_lstm} that the results are similar for both GRUs and LSTMs, with fine-tuning on Indian data. Also, all the models seem to improve with the fine-tuning on Indian data as depicted in Tables \ref{tab:results_finetune} and \ref{tab:results_lstm} with respect to Table \ref{tab:results_nofinetune}. As expected, in transfer learning, if a model pre-trained on data from the the source domain is fine-tuned on data from the transfer domain, the results seem to improve with respect to the model just pre-trained on the data from source domain \cite{yang2020}. 

Since GRUs and LSTMs yield similar results as seen in Tables \ref{tab:results_finetune} and \ref{tab:results_lstm} , GRUs have been preferred over LSTMs for the present experimentation as they have lesser number of parameters with respect to LSTMs \cite{ARUNKUMAR20227585}. Hence, for the rest of this discussion, results obtained only with the GRU networks are analyzed further.

\subsection{New Cases}

For the single country models, Bangladesh model gives the best results, followed by Brazil, Spain and USA. The model built using the data from Bangladesh is able to predict the trend in Indian data in a very accurate way. For the combination of two-country models, the presence of Bangladesh (with better trend tracking behaviour for Indian data) influences the performance in a positive way. Spain-Bangladesh combination gives the best result with the R-MSE of 0.0013 and R-MAE of 0.0177, and yields the best result amongst all the combinations. For the combination of more than two-country models, the performance does not improve further than the Spain-Bangladesh model. Overall, the two-country combination of Spain-Bangladesh model gives the best performance.

\subsection{New Deaths}

For the single country models, Brazil model gives the best results, followed by Bangladesh, Spain and USA. Bangladesh is again one of the better models, similar to the case of new cases. For the combination of two-country models, Spain-Bangladesh model gives the best results with a R-MSE of 0.0021 and R-MAE of 0.0286. Addition of data from the other two countries is not able decrease this error any further.

\subsection{Active Cases}

Prediction of active cases gives the best results as compared to the other two parameters. For the single country models, Brazil gives the best results, followed by USA, Spain and Bangladesh. For all the other models, the results improve only marginally for the Spain-Bangladesh model with a R-MSE of 0.0009 and R-MAE of 0.017. Rest of the results are all similar with no further improvement in the error metric. 

\subsection{Analysis}

For the two-country models, the Spain-Bangladesh combination (highlighted by light gray in Table \ref{tab:results_finetune}), gives a lower R-MSE than Bangladesh alone. There is an improvement in the result when pre-trained Spain model is combined with the pre-trained Bangladesh model. 

Once the combination of Spain-Bangladesh model has been built, further addition of models built with the other two countries (Brazil, USA) data does not reduce the error any further. This may be due to the following facts: Brazil has a high number of cases and has similar geography like that of India, the infections spread happened later than that of India.  

Bangladesh has similar climatic conditions and people's behaviour like that of India. Also, the percentage of people vaccinated with respect to the total population is similar in India and Bangladesh. As a result infection trend in Bangladesh has a positive impact in predicting infection trend in India. It may also be noted that, both India and Spain were vigilant at the start of the pandemic and had imposed strict infection control measures like lockdowns, social distancing etc. One such study \cite{Asawa2020} also corroborated this and compared the spread of COVID-19 infection in Spain and India by analysing the policy implications using epidemiological and social media data. Spain was one of the early COVID-19 infected countries, which is already at the end of fourth wave of infections. Whereas, India is at the end of the third wave. Also the spread increases sharply and then falls rapidly in both India and Spain. Such similar characteristics might be responsible for the low error in predictions obtained for the models built with Spain data.

It is to be noted that, the India model shown in the Table \ref{tab:results_finetune} is one where the GRU model is built with India data and then trained on India data i.e., it does not involve any transfer learning. It can be clearly seen from Tables \ref{tab:results_finetune} and \ref{tab:results_nofinetune} that the transfer learnt models are better predictors of all the three COVID-19 parameters. Models built with support vector regression (SVR), both with polynomial and RBF kernels, are unable to predict the COVID-19 parameters with a good level of accuracy. Same is the case with ARIMA and Facebook Prophet \cite{taylor2018forecasting}. Different nature of infection spread in different waves is difficult to take into account for SVR and ARIMA based predictions. This is seen from the high errors of prediction obtained using these models. Predicting daily new deaths is especially very uncertain. Presence of comorbidity, age etc. play a significant role in new deaths.

Hence, the proposed method, with the advantage of transfer learning from Bangladesh and Spain data combined, is able to predict the number of daily new cases, daily new deaths and active cases with the least error.

\begin{figure}[]
  \includegraphics[width=\linewidth]{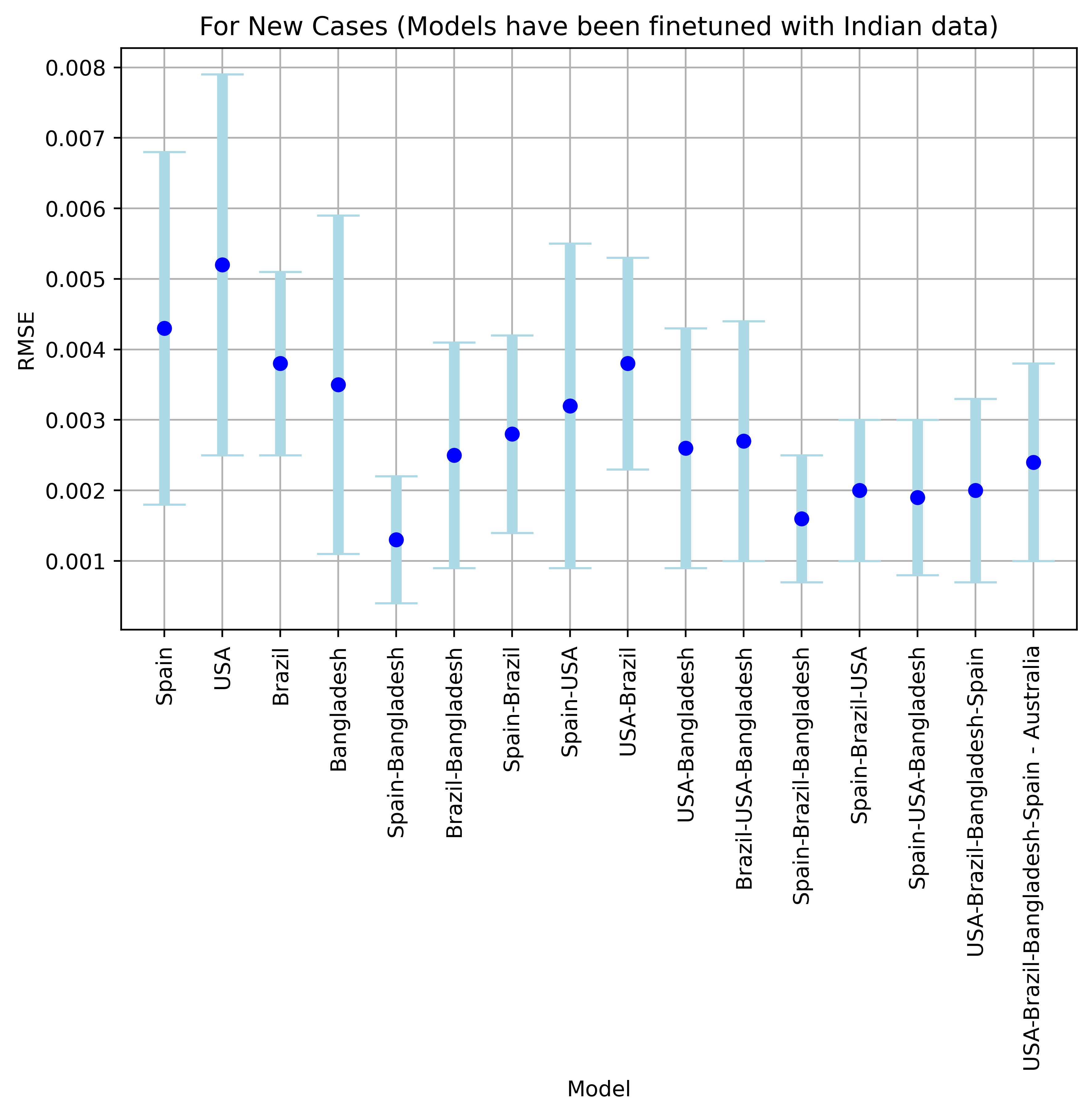}
  \caption{Average (over 20 runs) R-MSE after fine-tuning (blue dot) along with the standard deviation (light blue bars) for new cases prediction for all the models}
  \label{fig:rmse_std1}
\end{figure}

\begin{figure}[]
  \includegraphics[width=\linewidth]{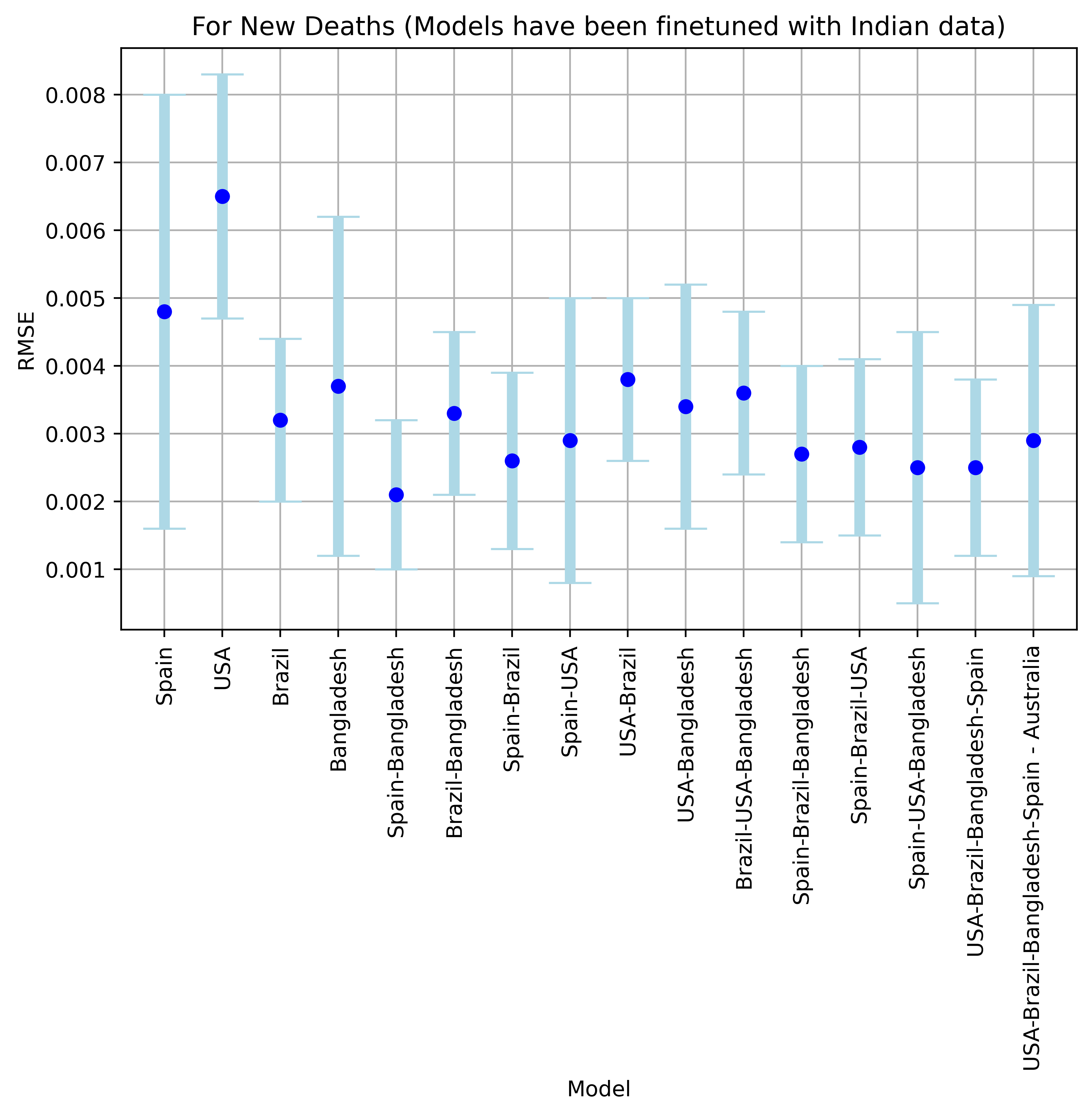}
  \caption{Average (over 20 runs) R-MSE after fine-tuning (blue dot) along with the standard deviation (light blue bars) for new deaths prediction for all the models}
  \label{fig:rmse_std2}
\end{figure}

\begin{figure}[]
  \includegraphics[width=\linewidth]{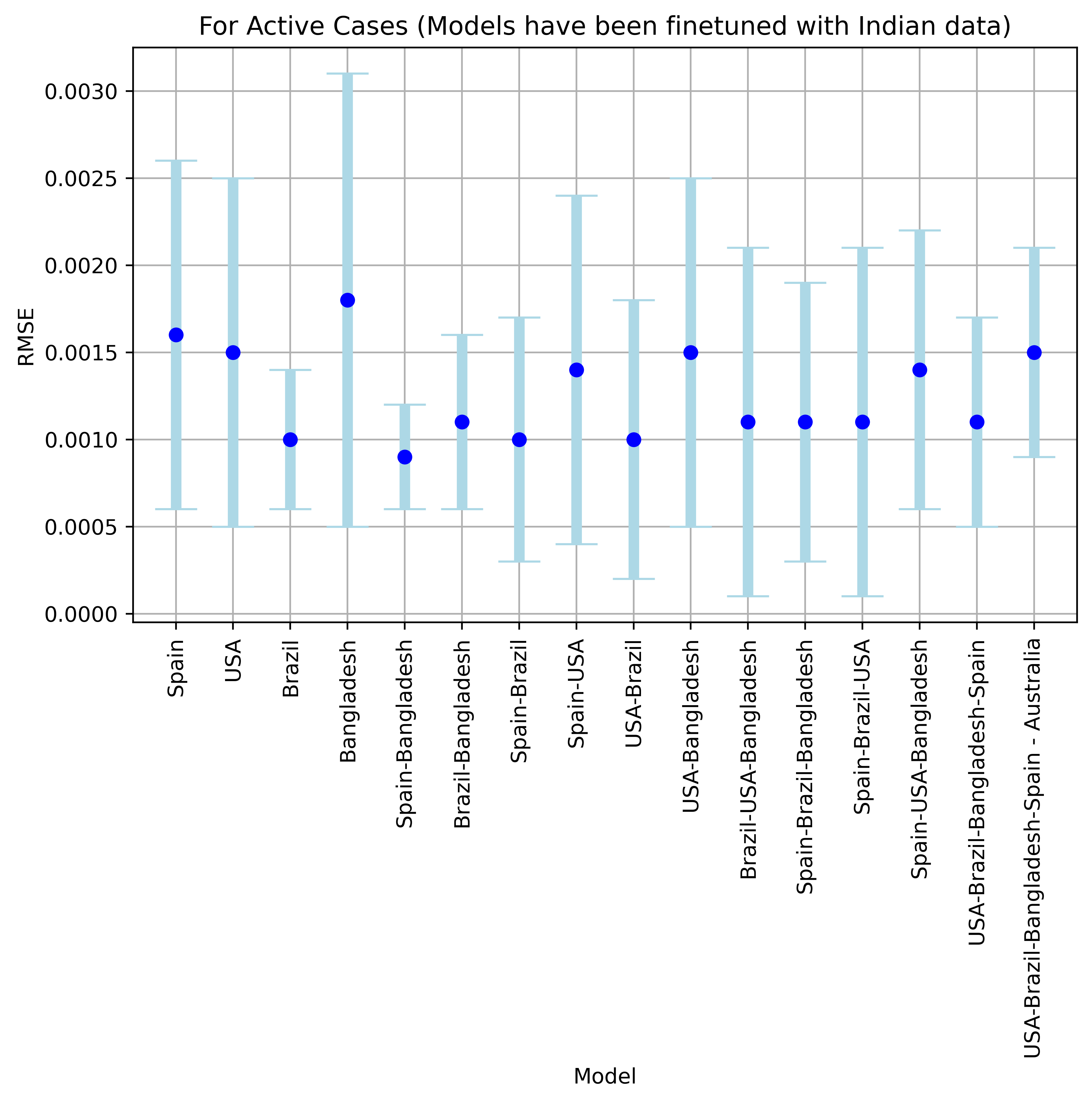}
  \caption{Average (over 20 runs) R-MSE after fine-tuning (blue dot) along with the standard deviation (light blue bars) for active cases prediction for all the models}
  \label{fig:rmse_std3}
\end{figure}

Standard deviation (STD) of the results are also studied. STD values for 20 runs for the prediction of new cases, for all the models fine-tuned on India data, are given in light blue in Figures \ref{fig:rmse_std1} to \ref{fig:rmse_std3}, showing clearly the Spain-Bangladesh combination to be the best model.

\begin{table*}[h]
\begin{tabular}{@{}lllllll@{}}
\toprule
\multicolumn{1}{c}{}                         & \multicolumn{2}{c}{New Cases} & \multicolumn{2}{c}{New Deaths} & \multicolumn{2}{c}{Active Cases} \\ \cmidrule(l){2-7} 
\multicolumn{1}{c}{\multirow{-2}{*}{Ensemble Type}} & R-MSE          & R-MAE          & R-MSE           & R-MAE          & R-MSE            & R-MAE           \\ \cmidrule(r){1-7}
Proposed Weighted Ensemble                   & 0.0013 & 0.0177 & 0.0021 & 0.0286 & 0.0009 & 0.017 \\
Equally Weighted Ensemble                     & 0.0049 & 0.0635 & 0.0055 & 0.0791 & 0.0021 & 0.0384 \\ \bottomrule
\end{tabular}
\caption{Comparison of results for Spain-Bangladesh model (with fine-tuning), using the proposed weighted ensemble and equally weighted ensemble}
\label{tab:equally_weighted_ensemble}
\end{table*}

\begin{figure}[]
  \includegraphics[width=\linewidth]{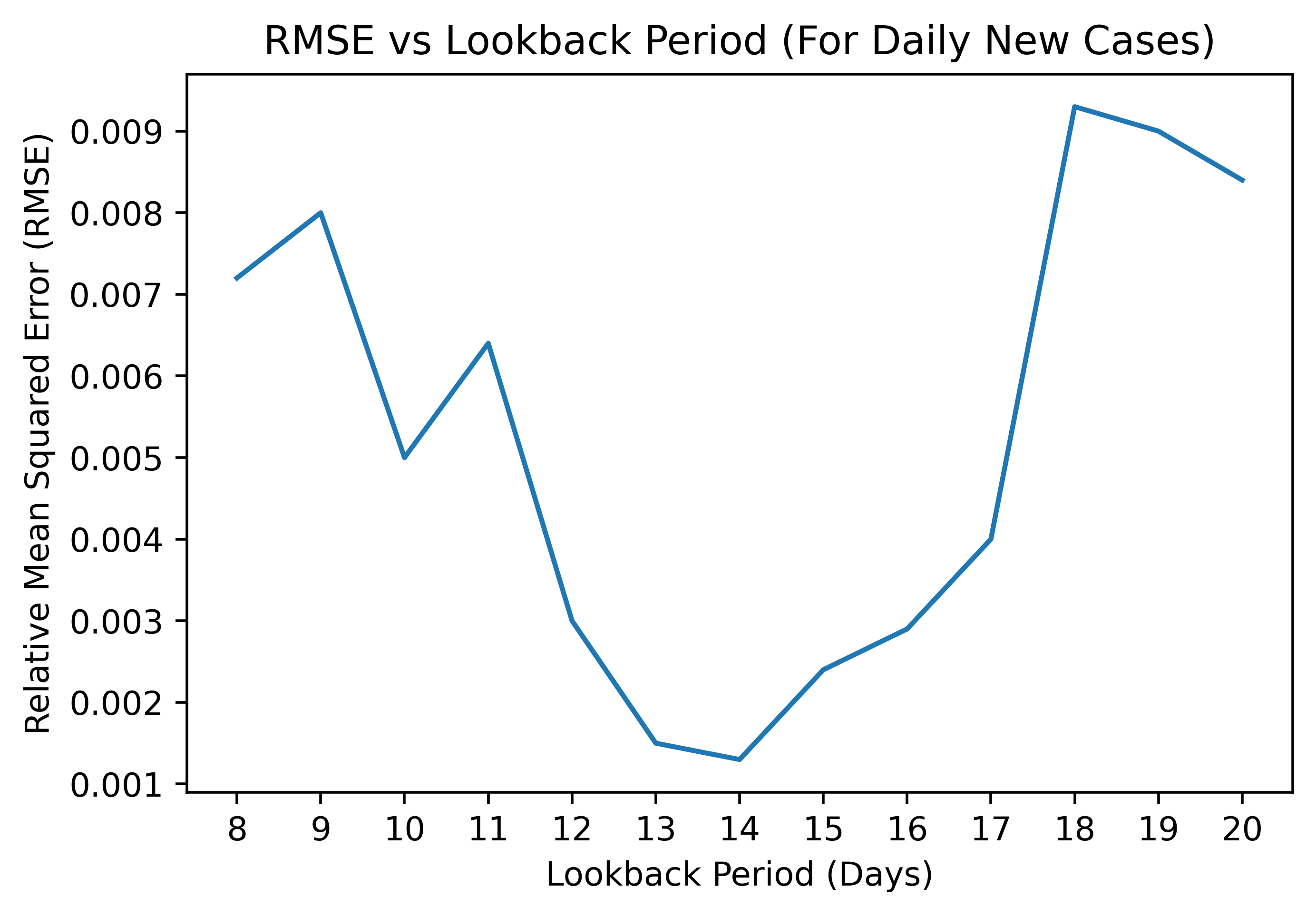}
  \caption{Prediction error for new cases vs Look-back period}
  \label{fig:lookback_newcases}
\end{figure}

\begin{figure}[]
  \includegraphics[width=\linewidth]{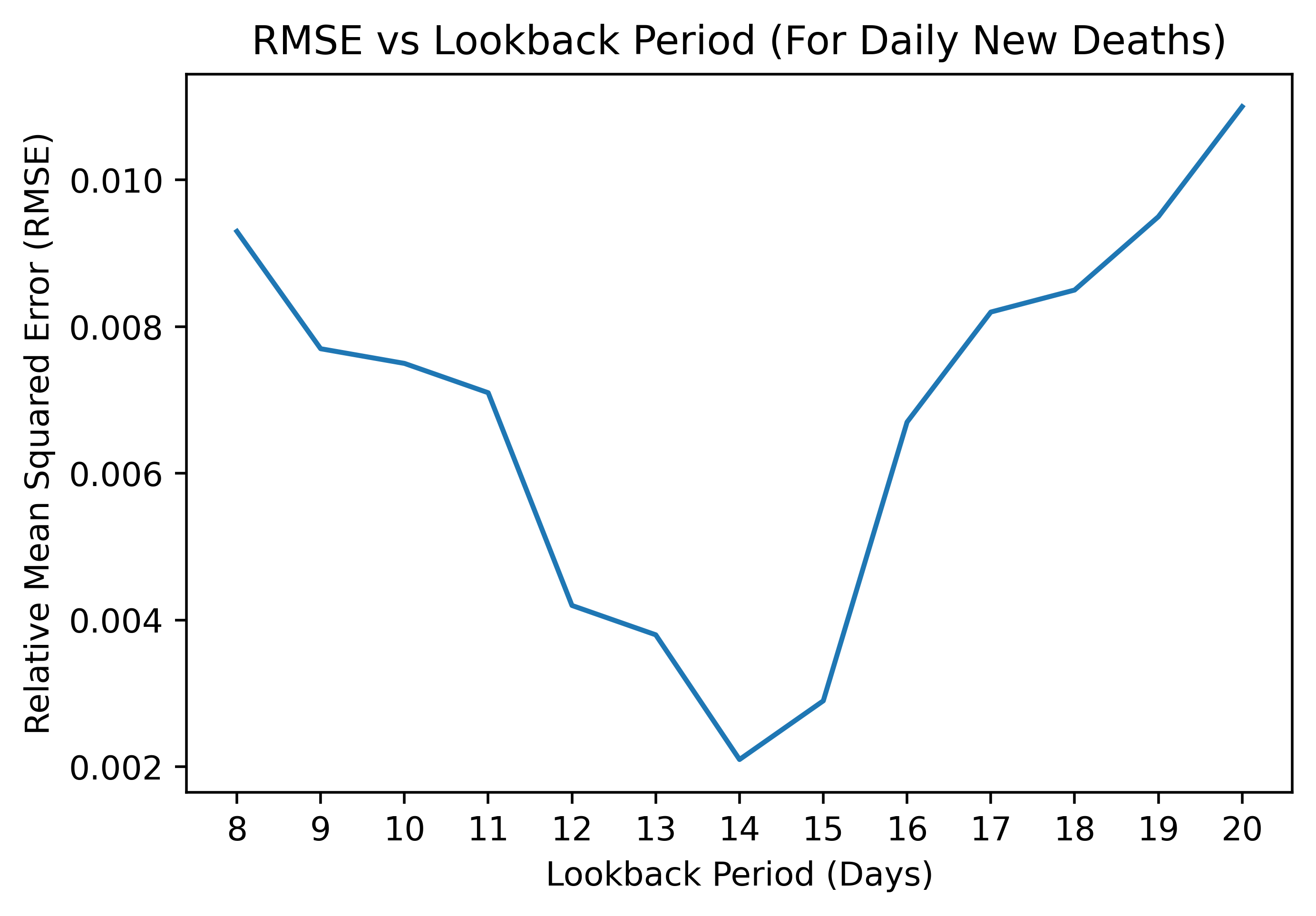}
  \caption{Prediction error for new deaths vs Look-back period}
  \label{fig:lookback_newdeaths}
\end{figure}

\begin{figure}[]
  \includegraphics[width=\linewidth]{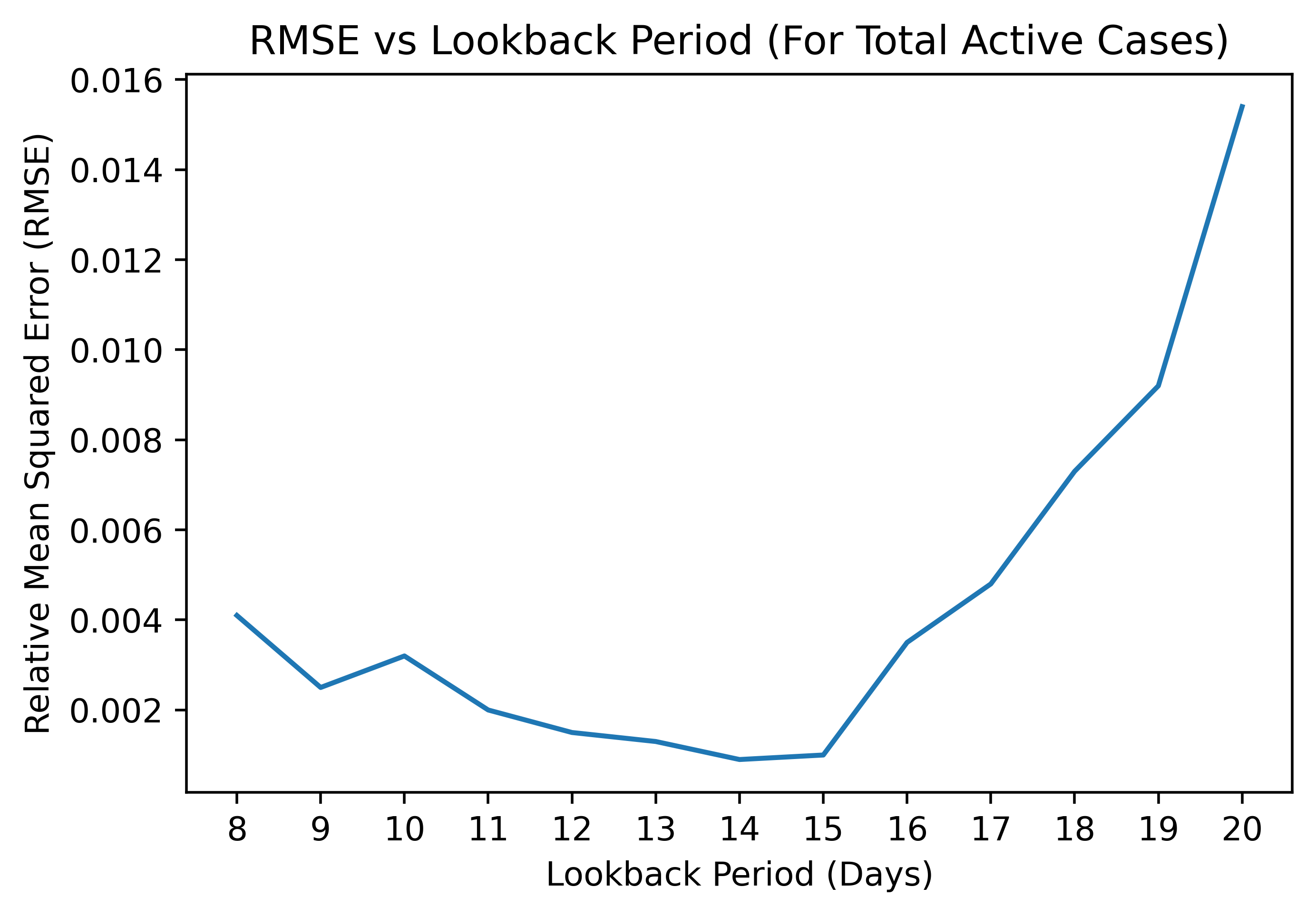}
  \caption{Prediction error for active cases vs Look-back period}
  \label{fig:lookback_activecases}
\end{figure}

The variation of prediction error with look-back period for all the three parameters is shown in Figures \ref{fig:lookback_newcases}, \ref{fig:lookback_newdeaths} and \ref{fig:lookback_activecases}. The prediction error is seen to be the least for the look-back period of 14 for all the three cases.

As mentioned earlier in the present method, a weighted ensemble is used for combining the models built with individual countries. Performance of the weighted ensemble method is compared with that of an equally weighted ensemble (where simple average is taken for combining the models) and the results for the best model (Spain-Bangladesh) is shown in Table \ref{tab:equally_weighted_ensemble}. It is seen that our proposed method with weighted ensemble performs better than the equally weighted ensemble approach.

\subsection{Statistical Significance Testing}

The proposed method was tested on a total of 167 test sets, with each set having a duration of 7 days (i.e., 7 day prediction). In order to do a statistical significance testing of the best model (Spain-Bangladesh model) obtained in our comparisons, it has been statistically tested against each of the other single country and two-country models using the Wilcoxon signed rank test. 

More than two country models were not used for calculating the statistical significance as they can be thought of as extensions of the two-country models. Only results for the prediction errors of new cases has been used in this testing.

Wilcoxon signed rank test is a non-parametric test for doing hypothesis testing of paired dependent samples. More details on Wilcoxon signed rank test can be found in \cite{woolson2007}.

For doing the statistical significance testing, 167 number of R-MSE values (for the 167 test sets) for two models  involved are treated as 167 number of paired observations. Wilcoxon signed rank test has been used over paired t-test as the difference between the observations are seen to be not normally distributed. Here,

\textbf{Null hypothesis - H$_0$}: There is no difference between the model and Spain-Bangladesh model

\textbf{Alternate hypothesis - H$_1$}: There is a difference between the model and Spain-Bangladesh model

\begin{table}[]
\begin{tabular}{@{}ll@{}}
\toprule
Model                & Z-Score \\ \midrule
Spain               & -9.59   \\
USA                 & -10.81  \\
Brazil              & -10.79  \\
Bangladesh          & -7.63   \\
Brazil - Bangladesh & -6.29   \\
Spain - Brazil      & -9.38   \\
Spain - USA         & -8.28   \\
USA - Brazil        & -10.85  \\
USA - Bangladesh    & -7.28   \\ \bottomrule
\end{tabular}
\caption{Wilcoxon signed rank test Z-score of the models with respect to Spain-Bangladesh Model}
\label{tab:stats_testing}
\end{table}

Two tailed hypothesis with a significance level of 0.05 has been used for the testing. The Z-score for each of the one-country and two-country models when compared with Spain-Bangladesh model is given in Table \ref{tab:stats_testing}. All the Z-scores in the Table \ref{tab:stats_testing} are less than -1.96 which is the critical Z-score for a two-tailed test at a significance level of 0.05. Hence, it can be concluded that the Spain-Bangladesh model is statistically significant from the other models in consideration.

\section{Conclusion and Future Work}

The proposed method with the combination of Spain-Bangladesh models outperforms the other combinations as well as other traditional regression models considered in our experiment. This is because the proposed method leverages the capabilities of both transfer learning and ensemble learning while taking into account the excellent sequence modelling capabilities of GRUs. The multi-day ahead prediction using recursive learning provides an added benefit of knowing the COVID-19 statistics multiple days ahead. The proposed method has currently been tested only on India data. Similar study and prediction can be done for other countries by choosing countries with transfer learning relevance. A regional study of COVID-19 cases is also necessary and the proposed method can be extended for individual states of India by incorporating information of other Indian states or other countries with comparable features with that of the individual Indian state. Individual waves of infection can also be analysed by using the transfer learning phenomenon from other waves of infection.

\section{Acknowledgments}
Support from the Science and Engineering Research Board (SERB) via sanctioning an ASEAN-India collaborative research project "A Multi Modal Approach to Medical Diagnosis Embedding Deep and Transfer Learning" (IMRC/AIST/DF/CRD/2019/000151 dated 29 June 2020) is gratefully acknowledged.

\bibliographystyle{cas-model2-names}
\bibliography{cas-refs}

\end{document}